\documentclass[10pt,twocolumn,letterpaper]{article}

\usepackage{iccv}
\usepackage{times}
\usepackage{epsfig}
\usepackage{graphicx}
\usepackage{amsmath}
\usepackage{amssymb}
\usepackage{pifont}
\usepackage{booktabs}
\usepackage[sort,nocompress]{cite}
\newcommand{\xmark}{\ding{55}}%
\usepackage[table,x11names]{xcolor}
\usepackage{soul}

\iccvfinalcopy

\usepackage{float}
\restylefloat{table}

\definecolor{aliceblue}{rgb}{0.94, 0.97, 1.0}
\definecolor{mistyrose}{rgb}{1.0, 0.89, 0.88}
\DeclareRobustCommand{\MA}[1]{{\sethlcolor{aliceblue}\hl{#1}}}
\DeclareRobustCommand{\rosehl}[1]{{\sethlcolor{mistyrose}\hl{#1}}}

\def\model{CATE\xspace}

\usepackage[pagebackref=true,breaklinks=true,letterpaper=true,colorlinks,bookmarks=false]{hyperref}
\newcommand{\var}{\mathbf}

\ificcvfinal\fi

\begin{document}

\title{Composable Augmentation Encoding for Video Representation Learning}

\author{Chen Sun$^{1,2}$
\quad
Arsha Nagrani$^{1}$
\quad
Yonglong Tian$^{3}$
\quad
Cordelia Schmid$^{1}$ \\
$^{1}$ Google Research \quad $^{2}$ Brown University \quad $^{3}$ MIT \\
{\tt\small \{chensun, anagrani, cordelias\}@google.com \quad yonglong@mit.edu}
}

\maketitle
\ificcvfinal\thispagestyle{empty}\fi
\begin{abstract}

We focus on contrastive methods for self-supervised video representation learning. A common paradigm in contrastive learning is to construct positive pairs by sampling different data views for the same instance, with different data instances as negatives. These methods implicitly assume a set of representational invariances to the view selection mechanism (\eg, sampling frames with temporal shifts), which may lead to poor performance on downstream tasks which violate these invariances (fine-grained video action recognition that would benefit from temporal information). To overcome this limitation, we propose an `augmentation aware' contrastive learning framework, where we explicitly provide a sequence of augmentation parameterisations (such as the values of the time shifts used to create data views) as composable augmentation encodings (\model) to our model when projecting the video representations for contrastive learning. We show that representations learned by our method encode valuable information about specified spatial or temporal augmentation, and in doing so also achieve state-of-the-art performance on a number of video benchmarks.
\end{abstract}

\section{Introduction}

\begin{figure}[t]
\centering
\includegraphics[width=\columnwidth]{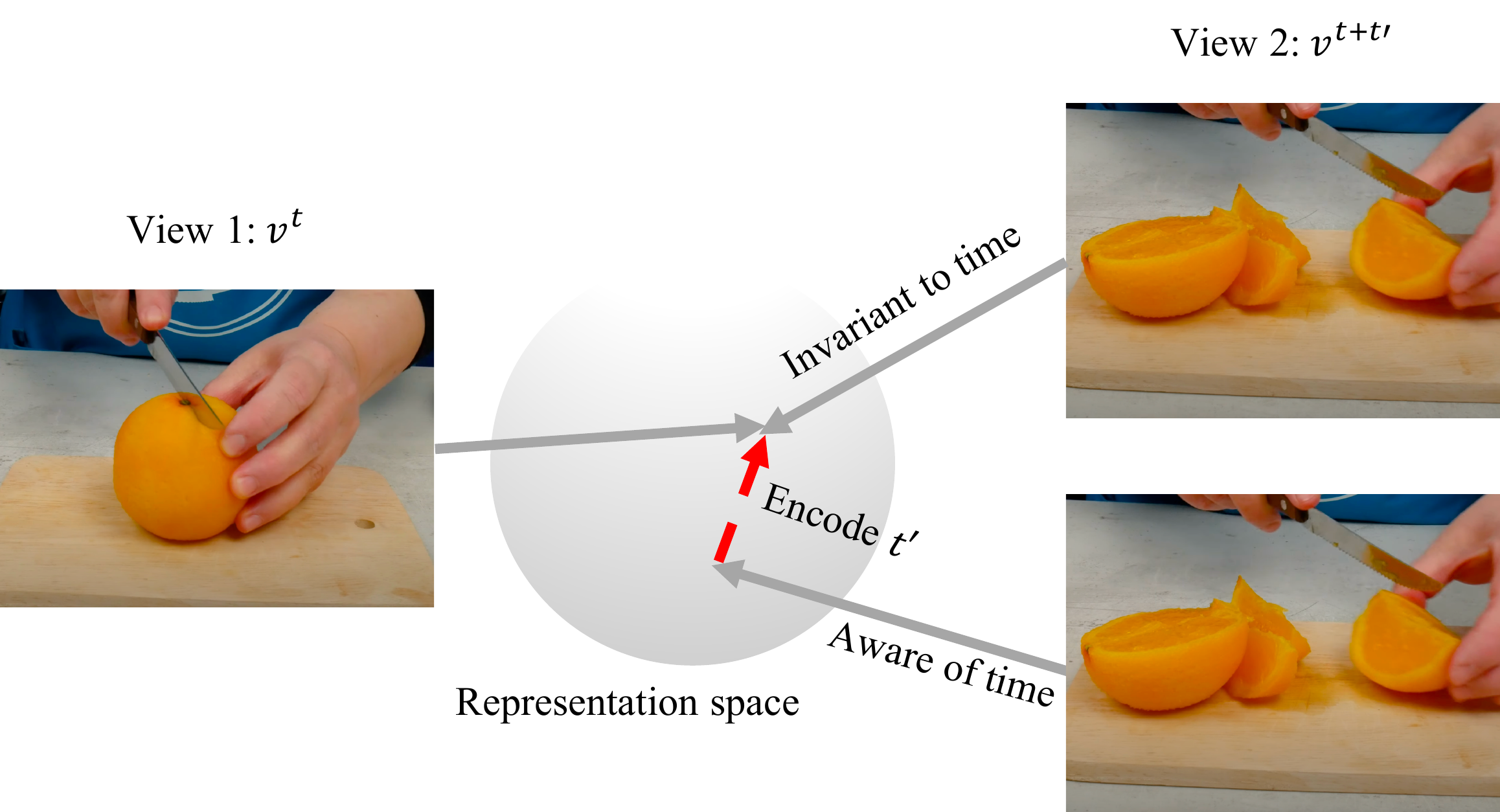}
\caption{Standard self-supervised contrastive learning methods are \textit{augmentation-invariant}, they ingest two augmented views of the same instance and encourage their latent representations to be similar.
For videos, if the two views were sampled with a temporal shift, this approach
would learn representations that are invariant to time changes, losing valuable temporal dynamics information (\eg the action of slicing oranges) for downstream tasks. We instead project latent representations \textit{by additionally encoding} the \textit{relative transformation} of the two views (\eg the time shift by $t'$) with a composable augmentation encoding (\model) to make the representation \textit{augmentation-aware}.
}
\label{fig:teaser} 
\end{figure}

We focus on contrastive learning~\cite{hadsell2006} of self-supervised video representations. The contrastive objective is simple: it pulls the latent representations of positive pairs close to each other, while pushing negative pairs apart. It is a natural fit for self-supervised learning, where positive and negative pairs can be constructed from the data itself, without the need for additional annotations.

Amongst different positive pair generation techniques, a particularly successful one has been augmentation-invariant contrastive learning~\cite{chen2020simple,he2019momentum,wu2018unsupervised}, which has shown impressive results for image representation learning. In this instance discrimination framework, positive pairs are constructed by applying artificial aggressive \textit{photo-geometric} data augmentations to create different versions of the same instances -- learned representations are thus encouraged to be invariant to these data augmentations. 

A number of self-supervised works also extend this idea to videos, where instead of artificial augmentations, they treat temporal shift as a \textit{natural} data augmentation~\cite{purushwalkam2020demystifying, tschannen2020self}, where two frames from the same video with a temporal shift form a positive pair. While this is useful for capturing high-level semantic information, \eg knowledge of object categories,
it can remove fine-grained information that may be useful depending on the downstream task -- as previously studied in images~\cite{chen2020simple,xiao2020should}. Consider the case of the oranges in Figure~\ref{fig:teaser} -- for object classification, invariance to small time shifts in a video that can result in view point changes or shape deformations may be useful and even desirable. However, for a downstream task involving reasoning about temporal relationships, such as recognising the state transition between a whole orange and orange slices caused by the act of cutting fruit, a representation invariant to temporal shifts may have lost valuable information.

We argue that augmentation-aware information can be retained if the relative augmentations of the two views are known to the contrastive learning framework. For the oranges in Figure~\ref{fig:teaser}, an encoder could keep the shape information of the sliced orange view if it is aware that the other view is $t'$ behind in time (and likely to be a whole orange).

We therefore propose a generalised framework for self-supervised video representation learning as follows. For notational convenience, we use the term \textit{data augmentation} to consider all parameterisable data transformations, including shifts in space or in time. We then apply these generalised data augmentations to create different views of the same data, as is done in previous augmentation-invariant contrastive learning. However, instead of directly applying a contrastive loss on these views, we apply an additional projection head that optionally also \textit{encodes the augmentations} that were used to create the views in the first place. For example, given two views of an image obtained via cropping, this can be an encoding of the bounding box co-ordinates of the cropping spatial transformation. For videos, this encoding can also include information about the temporal relationship between the views (\eg a 5-second shift), in addition to spatial transformations. In the case of missing or occluded sequential data, we can also specify the particular locations to be predicted. When no such encodings are provided, our framework becomes standard augmentation-invariant contrastive learning.

We formulate this framework as a prediction task, given sequential data as input. The input sequence in this case contains encoded visual representations to be learned, and optionally a set of encoded data transformations for prediction. By using transformers, we can easily compose multiple encoded data transformations in the input sequence. We refer to this projection as a \textbf{C}omposable \textbf{A}ugmen\textbf{T}ation \textbf{E}ncoding (\model) model. When data transformations are explicitly encoded in this way, our training objective can motivate a model to utilise such information if it helps with learning (\eg learning the temporal dynamics that oranges can be cut into slices). We choose to always sample negative pairs across different instances, so the model has the freedom to ignore the transformation encodings if they do not help in reducing the contrastive loss. However, empirical results show that they are almost always utilised.
We conduct thorough evaluations to test the efficacy of our framework, and in doing so make the following contributions: (1) we propose  Composable AugmenTation Encoding (\model) to learn augmentation-aware representations, and validate that \model learns representations that preserve useful information (\eg location, arrow of time) more effectively than a view-invariant baseline without augmentation encoding; (2) we perform a number of ablations on augmentation type and parameterisation, and observe that different downstream tasks favour the awareness of different augmentations, \eg temporal awareness is particularly helpful for fine-grained action recognition. We also find that encoding both the arrow of time and the absolute value of the temporal shift outperforms using just the arrow of time while encoding temporal information; (3) we set a new state-of-the-art for self-supervised learning on Something-Something~\cite{goyal2017something}, a dataset designed for fine-grained action recognition,
and finally (4) we also achieve state-of-the-art performance on standard benchmarks such as HMDB51~\cite{kuehne2011} and UCF101~\cite{ucf101}.

\section{Related Works} 
\noindent\textbf{Contrastive Learning.} Recently the most competitive self-supervised representation learning methods have used contrastive learning~\cite{oord2018cpc,wu2018unsupervised,hjelm2018learning,henaff2019data,tian2019contrastive,he2019momentum,chen2020simple,chen2020improved}. This idea dates back to~\cite{hadsell2006}, where contrastive learning was formulated as binary classification with margins. Modern contrastive approaches rely on a large number of negatives~\cite{wu2018unsupervised,tian2019contrastive,he2019momentum} and therefore a common technique is to employ the k-pair InfoNCE loss~\cite{sohn2016improved,oord2018cpc}. By choosing different positive pairs or `view' of the data, contrastive learning can encourage different representational invariances, e.g., luminance and chrominance~\cite{tian2019contrastive}, rotation~\cite{misra2019self}
, image augmentations~\cite{he2019momentum,chen2020simple}, temporal shifts~\cite{oord2018cpc,sermanet2018time,han2019dpc,zhuang2019unsupervised,qian2020spatiotemporal}, text and its contexts~\cite{mikolov2013distributed,logeswaran2018efficient,Kong2020A}, and multi-modal invariance~\cite{morgado2020audio,chung2019perfect,patrick2020multi,radford2021learning}. 

A number of works have highlighted the issues with these inbuilt representational invariances. InfoMin~\cite{tian2020makes} demonstrates that different invariances favour different downstream tasks, and proposes that optimal views for a given task should only be invariant to irrelevant factors of that task. In~\cite{purushwalkam2020demystifying}, occlusion-invariance is shown to benefit the downstream task of object detection. However, designing task-dependent invariances requires knowing the downstream task beforehand and may make the learned representations less general. To overcome this,~\cite{xiao2020should} learns multiple embedding spaces, each invariant to all but one augmentation. This means that the model (projection head) complexity grows linearly with the cardinality of the set of invariances. Instead, our transformer based projection head takes in a sequence of composable encodings, and can modulate the invariances that we want with a fixed model complexity. 

\begin{figure*}[t]
\centering
\includegraphics[width=.99\linewidth]{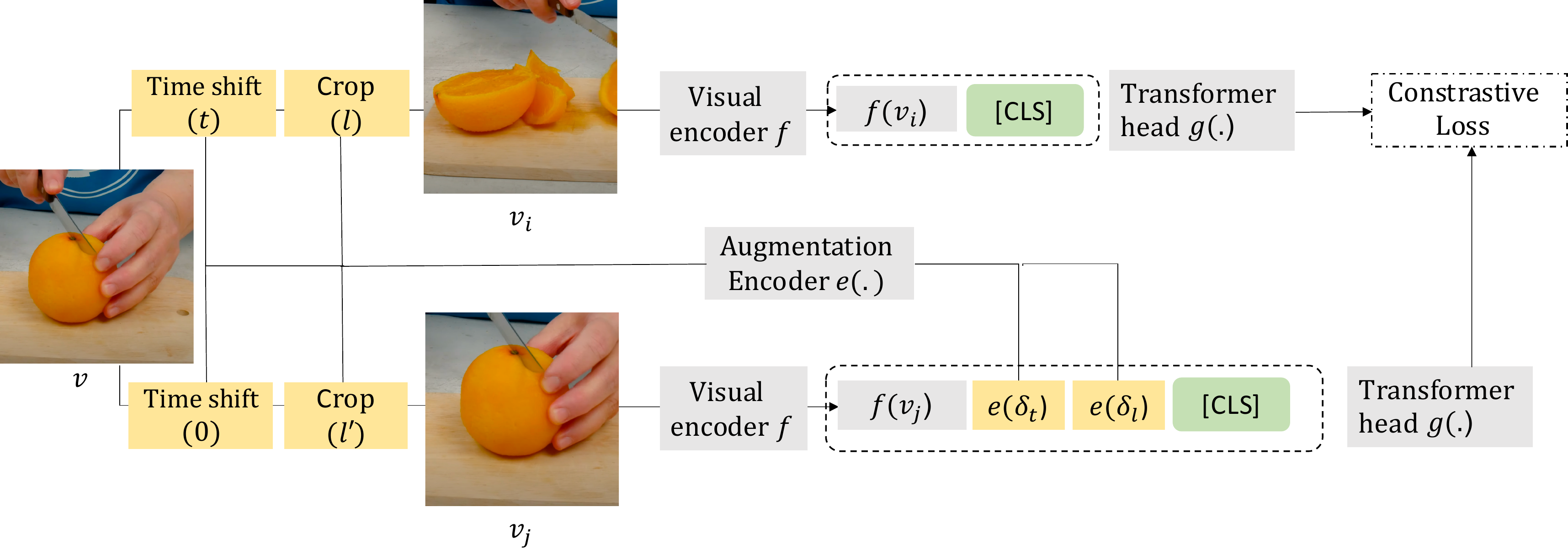}
\caption{\textbf{Overview of our contrastive learning framework \model:} Positive pairs are constructed from the same instance. For each view, a random set of data augmentations (e.g.\ temporal shifting, spatial cropping) is sampled and applied. The views are then encoded by a shared visual encoder (\eg 3D ConvNets for videos). Encoded visual features, along with parameterised and embedded data augmentations, are then passed to a transformer head (this contains multiple layers, only the input layer is shown for simplicity) which summarises the input sequence and generates projected features for contrastive learning. In this example, the bottom transformer head is tasked to predict the features knowing the temporal augmentation (predict features $t$ seconds ahead in time) and spatial augmentation (shift of box coordinates) relative to $v_i$.
The visual encoder $f$ is transferred to the downstream tasks.}
\label{fig:flow} 
\end{figure*}

\noindent\textbf{Self-supervised Learning for Video.}
Videos offer additional opportunities for learning representations beyond those of images, by exploiting spatio-temporal and multimodal~\cite{arandjelovic2017look,owens2016ambient} information. Some works extend spatial tasks to the space-time dimensions of videos \eg by using rotation~\cite{jing2018self} or jigsaw solving~\cite{kim2019self}, while others use temporal information by ordering frames or clips~\cite{OPN,fernando2017self,Misra2016ShuffleAL,wei2018learning,Xu_2019_CVPR,luo2020video,jenni2020video}, future prediction~\cite{vondrick2016anticipating,han2019dpc,han2020memory}, speed prediction~\cite{benaim2020speednet,wang2020self}, or motion~\cite{agrawal2015learning,diba2019dynamonet}. 
TaCo~\cite{bai2020taco} consolidates these works by combining different temporal augmentations.
Similar to us, they try to build augmentation awareness, but do so by adding a different \textit{pretext task} head besides the projection head for each temporally transformed video, which once again grows linearly with the cardinal set of augmentations.

\section{Method}

This section describes our unified \model framework for contrastive learning. We begin by providing an overview of contrastive learning. We then describe two paradigms in contrastive learning - `view-invariant' and `predictive coding', and show how our framework consolidates both using a transformer projection head. We also discuss how existing contrastive learning approaches can be viewed as special cases of our framework. An illustrative overview of our proposed framework is provided in Figure~\ref{fig:flow}. 

\subsection{Contrastive Learning}

Contrastive learning methods learn representations by maximising agreement between differently augmented
views of the same data example (a positive pair) while pushing apart different data samples (negative pairs). The construction of views for a positive pair can be very general, \eg,  via a stochastic data augmentation module or by sampling co-occurring modalities from multi-sensory data. Formally, given two random variables $\var{x}$ and $\var{y}$ -- contrastive learning seeks to learn a function that discriminates samples from the joint distribution $p(\var{x}, \var{y})$ and samples from the product of marginals $p(\var{x})p(\var{y})$. Therefore, there is a natural connection with mutual information maximisation, and as shown in~\cite{oord2018cpc} contrastive learning can be viewed as maximising a lower bound on the mutual information between representations of $\var{x}$ and $\var{y}$. Specifically, given an anchor point $\var{x}$, its paired $\var{y^+}$, and a set of negatives $\var{y_i^-}$ ($i=1,2,...,K$), the InfoNCE loss is defined as follows:
\begin{equation}\label{eq:infonce_loss}
    \mathcal{L}_\text{NCE} = - \mathbb{E}\left[\log \frac{e^{h(\var{x}, \var{y^+})}}{e^{h(\var{x}, \var{y^+})} + \sum_{i=1}^{K}e^{h(\var{x}, \var{y_i^-})}}\right]
\end{equation}
The critical function $h(\cdot, \cdot)$ typically consists of one or more backbone networks~\cite{chen2020simple,he2019momentum,tian2019contrastive}, projection heads~\cite{wu2018unsupervised,chen2020simple}, and a cosine similarity function. This function is optimised to assign a high score to positive pairs $(\var{x}, \var{y^+})$ and a low score to negative pairs $(\var{x}, \var{y_i^-})$. Minimising this InfoNCE loss is equivalent to maximising a lower bound $I_\text{NCE}$ on the mutual information between $\var{x}$ and $\var{y}$, denoted as $I(\var{x};\var{y})$: 
\begin{equation}\label{eq:infonce_bound}
    I(\var{x};\var{y}) \geq \log(K) - \mathcal{L}_\text{NCE} = I_\text{NCE}(\mathbf{x}; \mathbf{y})
\end{equation}

\subsection{Invariant and Predictive Coding}

For visual representation learning, positive pairs can be constructed from unlabeled data in a self-supervised fashion. Following the notation of~\cite{chen2020simple}, a popular approach is instance discrimination: where two views $\var{v}_i$, $\var{v}_j$ of the same instance $\var{v}$ are generated by applying independently sampled random data augmentations:
\begin{align*}
\var{v}_i = a(\var{v}; \tau_i) &\qquad
\var{v}_j = a(\var{v}; \tau_j) \\
\var{x} = g(f(\var{v}_i))&\qquad
\var{y^+} = g(f(\var{v}_j))
\end{align*}

where $a(\cdot)$ is the data augmentation operation parameterised by $\tau$, $f(\cdot)$ is the visual encoder usually implemented using ConvNets whose outputs are used for transfer learning, and $g(\cdot)$ is a projection head implemented by a Multilayer Perceptron (MLP). Note here that both $f(\cdot)$ and $g(\cdot)$ are shared across views. We refer to this approach henceforth as invariant coding.

Predictive coding, on the other hand provides an alternative approach, where certain regions within the same instance are masked out, such as different words in a sentence, different objects in an image or different frames in a video. This creates two different views - observed and `missing' regions of the instance $\var{v}_i$, $\var{v}_j$:
\begin{align*}
\var{v}_i &= m(\var{v}; \tau_i)\\
\var{v}_j &= m(\var{v}; \tau_j)
\end{align*}
where $m(\cdot)$ is the masking function, parameterised by $\tau$. Without loss of generality, we assume $\tau$ to be additive. This is a reasonable assumption for visual data which can be considered as sequential, and the indices (\eg timestamp for video, pixel location for image) to be masked are additive. 
The representations to be contrasted are then computed as:
\begin{align*}
\var{x} &= g(f(\var{v}_i); \var{0})\\
\var{y^+} &= g(f(\var{v}_j); \tau_j - \tau_i)
\end{align*}

Unlike with view-invariant coding, the projection head $g(\cdot)$ now also takes $\tau$ as input, \ie it is tasked to predict the representation conditioned on $\tau$ (\eg which could represent the time information used to generate the views -- `five seconds ahead'), and $\var{0}$ indicates the identity operation. In practice, $g(\cdot)$ can be easily implemented with a recurrent neural network as in~\cite{oord2018cpc} or a transformer with positional embeddings as in~\cite{devlin2018bert}.

We note that the predictive projection head $g(\cdot)$ does not necessarily need to take $\tau$ into account, assuming $f(\cdot)$ already encodes information invariant of $\tau$. However, arguably a much simpler task is to have $f(\cdot)$ encode the information useful for prediction and $g(\cdot)$ serve as the predictor (\eg generating new words for text, or learning the temporal dynamics for videos.).

\subsection{A Unified \model Framework}

Both $a(\cdot)$ and $m(\cdot)$ can be seen as atomic data operations that `transform' the data based on $\tau$, and multiple data operations can be chained together. We can therefore unify invariant and predictive coding using the following notation:
\begin{align*}
\var{v}_i = t(\var{v}; \var{\tau}_i) &\qquad
\var{v}_j = t(\var{v}; \var{\tau}_j) \\
\var{x} = g(f(\var{v}_i); \var{0})&\qquad
\var{y^+} = g(f(\var{v}_j); e(\var{\tau}_j - \var{\tau}_i))
\end{align*}
where $\var{\tau}_i = [\tau_i^1, \tau_i^2, ...]$ is a sequence of atomic data transformations, $t(\cdot)$ applies the sequence of data augmentations on the input data, and $e(\cdot)$ is an encoder that takes $\var{\tau}$ as input.

Given a sequence of atomic data operations, \textit{we can control which are view-invariant and which are predictive} with $e(\cdot)$: when we instruct $e(\cdot)$ to ignore certain type of operations, our method is `view-invariant' to such operations, otherwise we claim that such operations are predictive. 

We implement $g(\cdot)$ using a  transformer. Our transformer model takes a set of inputs, among them encoded visual features $f(\var{v}_i)$. We project the features to have the same size as the hidden size of the transformer with a linear layer. Additionally, each selected data operation is encoded into an embedding by an encoder $e_i(\tau_i)$ dedicated to the operation type. For example, for cropping, the inputs would be the differences between the coordinates of cropped boxes (for implementation details see Sec. \ref{sec:expsetup}). Finally, we have a special {\tt\small [CLS]} token which `summarises' the information from visual features and augmentation embeddings, and outputs a single embedding for each instance. The output embedding is then projected with a linear layer to the deisred output size. A big advantage of our framework is that the transformer projection head can elegantly deal with variable length inputs, and hence multiple augmentation encodings can be \textit{composed} with a fixed model capacity. We demonstrate this in the experiments (Sec.~\ref{sec:ablation}) that composing both crop and time encoding improves performance.

Note that other common contrastive based learning techniques can be expressed as special cases of our generalised framework: if $e(\cdot)$ is set to always return $\var{0}$ (the identity), our model is akin to SimCLR~\cite{chen2020simple} using a transformer-based projection head. When the inputs $\tau$ to $e(\cdot)$ correspond to the masked indices from a sequence, our formulation coincides with that of BERT~\cite{devlin2018bert} or GPT~\cite{gpt3}.

\section{Experiments} 
We first describe the datasets and their experimental setup (Sec. \ref{sec:data}), and then delve into the implementation details  (Sec. \ref{sec:expsetup}) of our framework. We then investigate a number of model ablations to better understand the design choices of \model, described in Sec. \ref{sec:ablation}. To further analyse our framework, we also design and evaluate on two proxy task that require knowledge of time in video - predicting time shifts and early action classification (Sec. \ref{sec:proxy-task}). Finally, we compare performance to the state of the art on regular action classification benchmarks.  

\subsection{Datasets}
\label{sec:data}

\noindent\textbf{Something-something~\cite{goyal2017something} v1:} This is a video dataset focused on human object interactions. The dataset has 108,499 videos with 174 categories of fine-grained human-object interactions. The action categories are designed to be focused on temporal information - time is needed to  distinguish between \textit{picking up something} and \textit{putting down something}, and hence it has been observed~\cite{s3dg_2017} that for this dataset, capturing subtle temporal changes is important for good performance. We use Something-Something v1 (SSv1) for our main ablation experiments. \\
\noindent\textbf{Something-something~\cite{goyal2017something} v2:} SSv2 is built on top of SSv1 by expanding the dataset size to 220,847 videos. The dataset is then augmented by~\cite{CVPR2020_SomethingElse} with object bounding box annotations. We use SSv2 to compare with previously published results. For both SSv1 and SSv2, we adopt the linear evaluation protocol used by self-supervised learning on ImageNet, where both pretraining and evaluation are done on the same dataset. At pre-training, videos in the training split are used to learn the representations. At the evaluation stage, we first train a supervised linear classifier on top of the frozen representations in the training split, then report classification accuracy on the validation set. To compare with others, we report results on Something-Else, which defines a split of SSv2 for few shot classification.\\
\noindent\textbf{Kinetics-400:} This dataset consists of 240K 10-second clips from YouTube videos with action labels covering 400 classes. We follow the standard practice of training on the trimmed clips in the training split but ignoring their action labels.
To evaluate representation learned on Kinetics-400, we follow standard practice and report results on the following two datasets:\\
\noindent\textbf{HMDB51:} HMDB51~\cite{kuehne2011} contains 6,766 video clips from 51 action classes. Evaluation is
performed using average classification accuracy over three train/test splits from~\cite{idrees2017thumos}, each with 3,570 train and 1,530 test videos.\\
\noindent\textbf{UCF101}: UCF101~\cite{ucf101} contains 13K videos downloaded from YouTube spanning over 101 human action classes. Similar to HMDB51 and as is standardly done, evaluation is performed using average classification accuracy over three train/test splits.
We pretrain on Kinetics and then evaluate on HMDB51 and UC101 in the following two ways: \\ 
(i) \textbf{Standard action classification:} We report performance with both (a) linear evaluation on frozen features and (b) finetuning. This is to compare with the state-of-the-art. \\
(ii) \textbf{Early action classification:}
This is to further understand the value of \model, and results are provided in Sec. \ref{sec:proxy-task}. Here we predict high level actions that may be performed in the future, given noisy visual evidence at the current time stamp, similar to what has been explored previously for action detection~\cite{sun2019relational}.
For this task, we train on just the first frame of the video in UCF and HMDB, and use only this single frame at test time as well.
\begin{table*}[htbp]
\begin{minipage}{.5\linewidth}
\scalebox{0.85}{
  \centering
    \begin{tabular}{ccccc}
  \toprule
  Aug. Time & Enc. Time & Projection & Top-1 Acc. & Top-5 Acc. \\
  \midrule
  \rowcolor{mistyrose}\xmark & \xmark & MLP & 17.1 & 40.9 \\
  \checkmark & \xmark & Linear & 20.9 & 45.9 \\
  \checkmark & \xmark & MLP & 26.4 & 55.2 \\
  \rowcolor{aliceblue}\checkmark & \xmark & Transformer & 26.5 & 55.9 \\
  \checkmark & \xmark & Tr. + MLP & 24.6 & 53.2 \\
  \checkmark & \checkmark & Transformer & \textbf{31.2} & \textbf{61.4} \\
  \bottomrule
  \end{tabular}
  }
\end{minipage}
\qquad
\begin{minipage}{.45\linewidth}
\scalebox{0.85}{
  \centering
  \begin{tabular}{ccccc}
  \toprule
  Encoded& $\tau$ & Dropout & Top-1 Acc. & Top-5 Acc. \\
  \midrule
  No & - & - & 26.5 & 55.9\\
  Crop & $\delta_{x,y}$ & \xmark & 27.2 & 56.7 \\
  Crop & $\delta_{x,y}$ &\checkmark & 28.1 & 58.0 \\
  Time & $\mathrm{sgn}(\delta_t)$ & \xmark& 28.1 & 57.9 \\
  Time & $\delta_t$ & \xmark & 31.3 & 62.4 \\
  Time & $\delta_t$ &\checkmark & 31.2 & 61.4 \\
  \bottomrule
  \end{tabular}
  }
\end{minipage}

\vspace{0.3cm}
\caption{\textbf{Left: Ablation showing the value of encoding time with different transformer projection heads} and \textbf{Right: Effect of regularisation on encodings}. All results are on SSv1 with linear eval on frozen features. Top row: \rosehl{Vanilla simCLR} Row four: \protect\MA{SimCLR++, simCLR with time augmentation and transformer projection head}. Last Row: When we encode time, we get a large boost in performance. On the right we show the impact of adding dropout to two aug. encodings - crop and time shifts. For crop encoding, dropout makes the pre-training task harder (higher contrastive loss at convergence), and also improves downstream accuracy. For time encoding, it is important to encode not only the arrow of time, but also the relative distance. However, dropout regularisation does not help in this setting.
}
\label{tab:head_encoding_ablation}
\end{table*}

\subsection{Implementation Details} \label{sec:expsetup}
\noindent\textbf{Base Model:}
Our implementation is based on the SimCLR~\cite{chen2020simple} code. Unless otherwise mentioned, we use a standard 3D ResNet-50 following the architecture of the `slow' branch in the SlowFast Networks~\cite{feichtenhofer2019slowfast}. Global batch normalization is used during contrastive pre-training, and local batch normalization is used during transfer learning. We use the standard data augmentations used by SimCLR~\cite{chen2020simple}: random cropping, color jittering and Gaussian blur. For spatial cropping, we found it beneficial to limit the range of cropped area to $(0.16, 0.81)$ of the original image area. For videos, we use time shifting as an additional atomic data operation. All the above spatial augmentations are applied consistently over time to avoid corrupting temporal continuity.
We use a lightweight transformer encoder as the projection head, with a hidden size of 768 units, intermediate size of 3072 and number of attention heads 12. We use 4 transformer layers in total. We add a linear projection layer after the transformer with output size of 256. As investigated in Section~\ref{sec:ablation}, when used standalone with no encoded data augmentations, the transformer head performs comparably with the nonlinear projection head used by SimCLR. This allows us to focus on the impact of augmentation encoding. \\
\noindent\textbf{Augmentation Encoding:}
In this work we encode two augmentations - spatial cropping and temporal shift, as we found these to be most effective empirically for the downstream tasks of action classification. As noted by SimCLR, invariance to augmentations like color jittering and gaussian blur is beneficial for classification, and hence we do not encode these.
We note here that a method that could automatically select which augmentations to encode would be interesting, and we leave this for future work.
Augmentations are encoded by $e(\cdot)$ as follows: for cropping we record the 4 scalar values representing the bounding box ($x_1, y_1, h, w$) of the crop, we then compute the relative distance of the cropped boxes between two views $\delta_{x,y}$ and project it to 768-dim with a linear layer. For temporal shifting we encode the binary indicator for arrow of time, and then a single scalar representing the number of frames shifted by. Each is projected to 768-dim with an embedding lookup table respectively, then summed together. \\
\noindent\textbf{Pretraining:} We feed 16 input frames to the ResNet-50-3D backbone at pre-training, with a frame sampling stride of 4 for Kinetics, and 2 for SSv1 and SSv2. All frames are cropped and resized to $224\times 224$. The transformer projection head is jointly trained with the ResNet-3D backbone during pre-training. We use the LARS optimiser with an initial learning rate of 4.8 ($= 0.15 * \sqrt(\mbox{BatchSize})$), and weight decay of $10^{-4}$. Unless otherwise mentioned, we pre-train for 500 epochs, with a batch size of 1024.\\
\noindent\textbf{Evaluation:}
For linear evaluation, we freeze the pre-trained visual encoders $f(\cdot)$, extract the 2048-dim output features and train a linear classifier on top. The transformer projection head is not used at this stage. We sample 16 frames with a stride of 1 during training, and up to 8 sliding windows of 32 frames for evaluation, which covers the entire video span. This `multi-crop in time' evaluation protocol is standard practice used by prior work~\cite{feichtenhofer2019slowfast}. 
For linear evaluation, we use a momentum optimiser of learning rate 0.16 and batch size of 256. For finetuning, we lower the learning rate to 0.02 and reduce the batch size to 128. All models are trained for 50 epochs. 

\subsection{Model Ablations} 
\label{sec:ablation}
In this section we perform 5 ablations, all on the SSv1 dataset. We pretrain on SSv1 train set without labels and then train a single linear layer. We evaluate on the SSv1 validation set. We first ablate the two key design choices of \model: (1) the use of a transformer projection head compared to an MLP or linear layer (Table \ref{tab:head_encoding_ablation}, left), and (2) the way we parameterise and regularise augmentation encodings (Table \ref{tab:head_encoding_ablation}, right). In particular, we assess the impact of encoding both time shifts and their direction (arrow of time) (Table \ref{tab:head_encoding_ablation}, right). 
We then show (3) that it is possible to compose multiple augmentations in our framework, and finally we ablate some low level details such as (4) the number of layers in the transformer head and (5) the number of epochs used for training. 

\noindent\textbf{1. Different Projection Head Types.} In this section, we start with the vanilla SimCLR model, and then vary the following (i) adding in temporal augmentations while training our model, (ii) encoding time or not under our \model framework and (ii) whether we use a linear, MLP or transformer projection head. Results are shown in Table~\ref{tab:head_encoding_ablation} (left). \\
\noindent\textit{SimCLR:} Vanilla SimCLR~\cite{chen2020simple} is applied to video frames, with no temporal data augmentation. Spatial augmentations are applied on the same frames to create views, and an MLP projection head is used. Results on SSv1 can be seen in the first row of Table~\ref{tab:head_encoding_ablation} (left).\\
\noindent\textit{SimCLR++:} In addition to spatial augmentation, we sample frames at different times in the same video to create views (row 3). We then also replace the MLP projection head with a transformer projection head which takes only the encoded visual representation $f(\cdot)$. This baseline is shown in the fourth row of Table~\ref{tab:head_encoding_ablation} (left), highlighted in blue, and henceforth is referred to as SimCLR++. It is a strong baseline for us to compare to.
From Table~\ref{tab:head_encoding_ablation} (left), it is clear that temporal data augmentation is essential for video representation learning, leading to a 9\% gain on top-1 accuracy with an MLP projection head (rows 1 and 3). When no temporal augmentation is encoded, we observe that a nonlinear MLP projection head gives similar performance to the transformer projection head (rows 2 and 3), which validates that any performance improvements in \model are not due solely to direct replacing the MLP head with a transformer. We also observe that combining the transformer head with the MLP head leads to slightly worse performance (row 5).

\noindent\textbf{2. Encoding Augmentations.} We first observe in Table~\ref{tab:head_encoding_ablation} (left) that adding temporal encoding improves the top-1 accuracy by nearly 5\% from 26.5\% to 31.2\% over SimCLR++ (last row vs fourth row in blue).  

To further explore the efficacy of our augmentation encoding, in Table~\ref{tab:head_encoding_ablation} (right), we ablate on the augmentation type using two augmentations - cropping and temporal shifts (time). We also explore the method of parameterisation for temporal augmentation (using just the arrow of time $\mathrm{sgn}(\delta_t)$ or the distance in time $\delta_t$, which includes both the absolute value of the temporal shift and its direction), and also the regularisation on the encodings (dropout or no dropout). We observe that both crop encoding and time encoding on their own improve the classification accuracy over the baseline, with time encoding providing a slightly bigger boost. By comparing the fourth row and the fifth row, we can see that the parameterisation of the temporal augmentation also matters, and it is beneficial to pass the distance in time along with the arrow of time to the augmentation encoder. Finally, we observe that dropout regularisation helps crop encoding but not time encoding. We hypothesise that crop encoding might make the contrastive task too easy and stronger regularisation is needed; time encoding does not suffer from this issue as there is more variation to be learnt from different frames in a video, and it actually benefits from a more informative encoding (from $\mathrm{sgn}(\delta_t)$ to $\delta_t$).

\noindent\textbf{3. Composing multiple augmentations.} In Table~\ref{tab:compose_ssv1} we show results for composing both crop and time encodings before feeding them to the transformer head. We can see for that for SSv1 (similar results hold for SSv2 and can be found in Table~\ref{tab:compose_ssv2}, appendix), using crop and time encodings individually leads to improved performance over the no-encoding baseline (\ie SimCLR++, first row), and composing them together leads to further improvement. 
\begin{table}[h]\centering
\scalebox{0.8}{
  \begin{tabular}{cccc}
  \toprule
  Enc. Crop & Enc. Time & Top-1 Acc. & Top-5 Acc. \\
  \midrule
  \rowcolor{aliceblue}\xmark&\xmark&26.5&55.9\\
  \checkmark&\xmark&28.1&58.0\\
  \xmark&\checkmark&31.2&61.4\\
  \checkmark&\checkmark&\textbf{32.2}&\textbf{62.4}\\
  \bottomrule
  \end{tabular}
  }
\vspace{-0.1cm}
\caption{\textbf{Composing spatial (crop) and temporal encodings} for Something-Something v1. Each individual encoding outperforms the no encoding baseline (\protect\MA{SimCLR++}). Composing them together yields the best performance.}
\label{tab:compose_ssv1}
\end{table}

\noindent\textbf{4. Number of transformer layers.} We experiment with a number of Transformer layers (1,2,4,8) for our projection head (with time encoding), and observe that 
the performance begins to saturate at four layers (Table~\ref{tab:trn_ablation}, appendix). We use four layers in all other experiments.

\noindent\textbf{5. Number of training epochs.} We study the impact of the number of epochs used for pre-training on SSv1 and Kinetics-400. For evaluation we use SSv1, HMDB and UCF. Similar to SimCLR~\cite{chen2020simple}, we observe improved performance by increasing the number of epochs initially, which then saturates at around 500 epochs (Table~\ref{tab:something_epoch}, appendix).
\subsection{Further Analysis}\label{sec:proxy-task}
In this section we further analyse the effect of encoding augmentations on learned representations.\\ 

\noindent\textbf{Per-class breakdown on SSv1:}
We conjecture that augmentation encoding is helpful for the downstream tasks that need to be aware of the corresponding spatial and temporal augmentations. To verify this conjecture, in  Table~\ref{tab:time_per_class} and Table~\ref{tab:crop_per_class} (appendix) we list the SSv1 classes that benefit the most and the least from time and crop encoding, respectively. We sort the classes by computing their per-class Average Precision.
We can clearly see that the top classes for
time encoding are typically sensitive of temporal ordering by definition, such as \textit{lift up then drop down}, and \textit{move closer}, where changing the arrow of time would lead to the opposite action (\eg \textit{move farther away}). Similarly, for crop encoding, the classes that benefit the most are those that require some level of spatial reasoning (\eg \textit{lift up}, \textit{drop down}, \textit{pull from right to left}, and \textit{move down}), while the bottom classes typically do not require spatial reasoning.

\begin{table}[h]\centering
\scalebox{0.8}{
  \begin{tabular}{cc}
  \toprule
  Label & $\Delta\textrm{AP}$\\
  \midrule
  \rowcolor{aliceblue}Lifting something up completely, then letting it drop down & 21.0\\
  \rowcolor{aliceblue}Pulling two ends of something so that it gets stretched & 19.8\\
  \rowcolor{aliceblue}Moving something and something closer to each other & 18.5\\
  \rowcolor{aliceblue}Taking one of many similar things on the table & 17.2\\
  \rowcolor{aliceblue}Pushing something so that it almost falls off but doesn't & 16.7\\
  \rowcolor{mistyrose}Poking something so lightly that it doesn't move & -4.6\\
  \rowcolor{mistyrose}Pretending to pour something out of something & -5.4\\
  \rowcolor{mistyrose}Poking a stack of something without the stack collapsing & -5.5\\
  \rowcolor{mistyrose}Pretending to spread air onto something & -7.8\\
  \bottomrule
  \end{tabular}
 }
\vspace{-0.1em}
\caption{Classes that benefit the most and the least with \textbf{time encoding} on SSv1. We sort the classes by their differences on Average Precision.}
\label{tab:time_per_class}
\end{table}

\noindent\textbf{Predicting Time Shifts:} In previous experiments, we confirm empirically that encoding augmentations during pretraining leads to better downstream performance. As a sanity check however, we also further design a proxy task to verify that the representations are indeed storing the encoded information and not discarding it. To analyse the time encodings, we design a time shift classification experiment based on the SSv1 dataset. For each video, we sample two 16-frame clips and use their relative distance in time as the classification label. The label space is \textit{quantised} every 6 frames (0.5 seconds). During training and evaluation, we take the frozen representations of the two clips, concatenate them channel-wise and pass them to a linear classifier on top. Table~\ref{tab:something_aot} shows the results. We can see that by providing the encoded time augmentation $\delta_t$ during pre-training, \model learns representations that maintain temporal shift information, solving the task with near perfect accuracy. Providing only the arrow of time $\mathrm{sgn}(\delta_t)$ retains some information, while the no encoding baseline performs poorly on this probing task.

\begin{table}[h]\centering
\scalebox{0.8}{
  \begin{tabular}{cccc}
  \toprule
  Encode Time & $\tau$ & Time Offset Acc. \\
  \midrule
  \xmark & - & 5.7 \\
  \checkmark & $\mathrm{sgn}(\delta_t)$ & 65.7 \\
  \checkmark & $\delta_t$  & \textbf{99.9} \\
  \bottomrule
  \end{tabular}
} 
\vspace{-0.1cm}
\caption{\textbf{Time Shift Classification on SSv1}. Encoding time significantly helps on this proxy task, validating the intuition that our model retains useful time information.}
\label{tab:something_aot}
\end{table}

\noindent\textbf{Early Action Classification:} We use this benchmark to investigate the impact of the number of observed frames on action recognition. Table~\ref{tab:ablation_ucf_hmdb} in Section~\ref{supp:ablation} (Appendix) reports results on UCF-101 and HMDB-51. We observe that the gain of encoding time is bigger for early action classification than for full video classification, which indicates that time information is more important for the task.

\subsection{Comparison with State-of-the-Art}

Finally, we present comparisons with previous state-of-the-art methods on SS, UCF101 and HMDB51. 

For SS, we compare our self-supervised representations with other weakly- and fully-supervised representations. For evaluation, all representations are frozen, and a linear classifier is trained on the labeled training examples from the target dataset.
In Table~\ref{tab:something_sota}, we compare \model with competitive weakly-supervised methods. \model is pretrained on the train split of SSv1, and all weakly-supervised representations are pretrained by the authors of~\cite{yan2020clusterfit} on 19M public videos with hashtag supervision. The target dataset is SSv1.
Despite training with only 0.1M videos without using their labels, our method is able to outperform these weakly-supervised approaches by large margins.

\begin{table}[h]\centering
\scalebox{0.85}{
  \begin{tabular}{ccc}
  \toprule
  Method & Supervision & Top-1 Acc. \\
  \midrule
  \rowcolor{aliceblue}Distillation~\cite{hinton2015distilling} & Weak & 15.6\\
  \rowcolor{aliceblue}Prototype~\cite{prototype} & Weak & 20.3 \\
  \rowcolor{aliceblue}ClusterFit~\cite{yan2020clusterfit} & Weak & 20.6 \\
  SimCLR++~\cite{chen2020simple}$^*$ & Self & 26.4 \\
  \model & Self & \textbf{32.2}\\
\bottomrule
  \end{tabular}
  }
\vspace{0.1cm}
\caption{\textbf{Comparison to SoTA on the Something-something v1 val set}. We use linear evaluation on frozen features. We compare to weakly-supervised baselines by~\cite{yan2020clusterfit}. *: re-implemented with temporal augmentation.}
\vspace{-0.2cm}
\label{tab:something_sota}
\end{table}

In Table~\ref{tab:something_else}, we compare with fully-supervised Spatial-Temporal Interaction Networks (STIN)~\cite{CVPR2020_SomethingElse} for few-shot action classification. Both \model and STIN are pretrained on the `Base' split of Something-Else, which contains half of the videos. STIN uses its labels as supervision while \model does not. The target few-shot dataset contains 5, or 10 examples per class, across 86 classes. This is a more challenging setup than the 5-way classification setup used by~\cite{cao2020few}. We found \model achieves on par or better performance than the supervised STIN.

\begin{table}[h]\centering
\scalebox{0.8}{
  \begin{tabular}{cccc}
  \toprule
  Method & Pretrain & 5-shot Acc. & 10-shot Acc. \\
  \midrule
  \rowcolor{aliceblue}STIN+OIE+NL~\cite{CVPR2020_SomethingElse} & Supervised & 17.7 & 20.7\\
  SimCLR++~\cite{chen2020simple}$^*$ & Self-sup. &14.4 & 19.8 \\
  \model & Self-sup. &\textbf{18.0} & \textbf{22.9} \\
\bottomrule
  \end{tabular}
}
\vspace{0.1cm}
\caption{\textbf{Comparison to SoTA on Something-Else, a split of Something Something-v2 for few shot classification}. *: our re-implementation with temporal augmentation.}
\vspace{-0.2cm}
\label{tab:something_else}
\end{table}

\begin{table}[h]\centering
\footnotesize{

  \begin{tabular}{cccccc}
  \toprule
  Method & Modalities & Dataset\!\! &Frozen &  \!\!UCF \!\!& \!\!HMDB\!\! \\
  \midrule
    Shuff\&Lrn*~\cite{Misra2016ShuffleAL} & V& UCF &  \checkmark & 26.5&12.6  \\
  3DRotNet~\cite{jing2018selfsupervised} &V& K400 &  \checkmark & 47.7 & 24.8 \\
  CBT~\cite{sun2019contrastive} &V& K600 & \checkmark & 54.0 & 29.5  \\
  MemDPC~\cite{han2020memory} &V& K400 &\checkmark & 54.1 & 30.5 \\
  TaCo~\cite{bai2020taco} & V & K400 & \checkmark & 59.6 & 26.7 \\

  \textbf{\model}  &V& K400  &\checkmark& \textbf{84.3} & \textbf{53.6}\\
 \rowcolor{aliceblue}MemDPC~\cite{han2020memory} &V+F& K400 &\checkmark & 58.5 & 33.6 \\
    \rowcolor{aliceblue}CoCLR~\cite{han2020self}& V+F & K400  &\checkmark& 74.5&46.1 \\
 \rowcolor{aliceblue} AVSlowFast~\cite{xiao2020audiovisual}&V+A& K400 & \checkmark& 77.4 & 44.1 \\

  \rowcolor{aliceblue}MIL-NCE~\cite{miech2020end} &V+T& HTM & \checkmark & 83.4&54.8 \\
  \rowcolor{aliceblue}XDC~\cite{alwassel2020self} & V+A& IG65M & \checkmark & 85.3 & 56.0 \\
  \rowcolor{aliceblue}ELO~\cite{piergiovanni2020evolving} & V+A& YT8M & \checkmark & -- & 64.5 \\

  \midrule 
  Shuff\&Lrn*~\cite{Misra2016ShuffleAL}&V & UCF & \xmark &50.2 & 18.1 \\
  CMC~\cite{tian2019contrastive}&V & UCF& \xmark & 59.1 & 26.7 \\
  OPN~\cite{OPN}&V & UCF&\xmark & 59.6 & 23.8 \\
  ClipOrder~\cite{Xu_2019_CVPR}&V & UCF &\xmark& 72.4 & 30.9 \\
  3DRotNet~\cite{jing2018selfsupervised}&V & K400 &\xmark & 66.0 & 37.1 \\
  DPC~\cite{han2019dpc}&V & K400 &\xmark &75.7 & 35.7 \\
  CBT~\cite{sun2019contrastive}&V & K600 &\xmark & 77.0 &  47.2 \\
  MemDPC~\cite{han2020memory} &V& K400 &\xmark & 78.1 & 41.2 \\
  SpeedNet~\cite{benaim2020speednet} &V &K400 &\xmark &81.1 & 48.8 \\
  VTHCL~\cite{yang2020video}& V & K400 &\xmark & 82.1 & 49.2 \\
  TaCo~\cite{bai2020taco} & V & K400 & \xmark & 85.1 & 51.6 \\

  \textbf{\model}  &V& K400  &\xmark& \textbf{88.4} & \textbf{61.9}\\
       \rowcolor{aliceblue}MemDPC~\cite{han2020memory} &V+F& K400 & \xmark & 86.1 & 54.5 \\
      \rowcolor{aliceblue}CoCLR~\cite{han2020self}&V+F & K400& \xmark &87.9 &  54.6 \\
\rowcolor{aliceblue}MIL-NCE~\cite{miech2020end} &V+T& HTM &\xmark  & 91.3&61.0 \\

  \rowcolor{aliceblue}ELO~\cite{piergiovanni2020evolving} & V+A& YT8M & \xmark & 93.8 & 67.4 \\
    \rowcolor{aliceblue}XDC~\cite{alwassel2020self} & V+A& IG65M & \xmark & 94.2 &  67.4 \\

  \bottomrule
  \end{tabular}
 
\vspace{-0.2cm}
\caption{\textbf{Comparison to the state of the art on UCF101~\cite{ucf101} and HMDB51~\cite{kuehne2011}.} *reimplemented by ~\cite{sun2019contrastive}. \textbf{Frozen $\checkmark$} means the pretrained representation is fixed and classified with a linear layer, while \textbf{\xmark} means all layers are finetuned end-to-end. \protect\MA{Rows highlighted in light blue use modalities beyond RGB frames as sources of supervision.} Modalities are \textbf{V:} RGB frames only, \textbf{T:} text from ASR, \textbf{F:} pre-extracted optical flow, \textbf{A:} audio. }
\label{tab:vid_results_action}
}
\end{table}

We also compare to the state-of-the-art on both HMDB51 and UCF101 in Table~\ref{tab:vid_results_action}. Using frozen features, our model outperforms all other works that pretrain using RGB frames only -- on UCF we even outperform a large number of works that use end-to-end finetuning. Additionally, our model outperforms AVSlowFast~\cite{xiao2020audiovisual} which uses additional supervision from audio, and both MemDPC~\cite{han2020memory} and the recently proposed CoCLR~\cite{han2020self}, which use additional information from pre-extracted optical flow. Our model also compares favourably with MIL-NCE, XDC and ELO that are trained on orders of magnitude more training data -- IG65M consists of 21 years of data (XDC), HTM, 15 years (MIL-NCE), and YouTube 8M, 13 years (ELO). In contrast, Kinetics400 contains only 28 days of video data. 

On finetuning we note that the gaps are smaller, however we still outperform all previously published works that use RGB frames only. We note that methods that use additional information from other modalities and train on orders more training data (MIL-NCE, XDC and ELO) are able to almost saturate performance on the UCF dataset.

\section{Conclusion} 
We propose a general framework for contrastive learning, that allows us to build augmentation awareness in video representations. Our method consists of an elegant transformer head to encode augmentation information in a composable manner, and achieves state-of-the-art results for video representation learning.
Future work will include evaluating on structured video understanding tasks and measuring the extent of equivariance learned by the representations.
{\small
\bibliographystyle{ieee_fullname}
\bibliography{egbib}
}

\clearpage

\section{Appendix}
\setcounter{table}{0}
\renewcommand{\thetable}{A\arabic{table}}
\setcounter{figure}{0}
\renewcommand{\thefigure}{A\arabic{figure}}

\begin{table*}[htp]\centering
  \begin{tabular}{cccc}
  \toprule
   Table No. & Pretrain Data (Unlabeled) & Target Data (Train) & Target Data (Eval) \\
  \midrule
  1,2,3,5 & SSv1 train split & SSv1 train split & SSv1 val split\\
  4,7 & Kinetics-400 train split & UCF/HMDB train splits & UCF/HMDB val splits\\
  6 & SElse `Base' train split & SElse `Novel' train split & SElse `Novel' val split\\ 
  \bottomrule
  \end{tabular}
\caption{Pretraining datasets for self-supervised representation learning with \model, and target datasets for linear evaluation for results reported in the main paper.}
\label{tab:transfer_setup}
\end{table*}

\subsection{Evaluation Protocol}
Our work is about self-supervised pretraining of video representations. For evaluation, we mostly perform transfer learning experiments following the standard linear evaluation protocol commonly used by recent self-supervised image representation learning approaches~\cite{oord2018cpc,chen2020simple}. Our video representation is first pretrained on unlabeled videos from a large pretraining dataset. 

We then transfer the self-supervised representations to the target dataset, by training a linear classifier on top of the frozen representations. This linear classifier is trained on labeled examples from the training split of the target dataset. Accuracy on the test split of the target dataset is used to measure the representation quality. We list the pretrain and target datasets used to generate the results in our main submission in Table~\ref{tab:transfer_setup}.

\subsection{More Model Ablations}
\label{supp:ablation}
\noindent\textbf{1. Number of transformer layers.} 
We vary the number of transformer layers (1,2,4,8) for our projection head, which receives encoded time augmentations as additional input. Performance on SSv1 linear evaluation can be found in Table~\ref{tab:trn_ablation}. We observe that the performance begins to slightly saturate at four layers.

\begin{table}[H]\centering
\scalebox{0.8}{
  \begin{tabular}{ccc}
  \toprule
   No. Layers & Top-1 Acc. & Top-5 Acc. \\
  \midrule
  1 & 26.9 & 56.2 \\
  2 & 30.0 & 60.3 \\
  4 & 31.2 & 61.4 \\
  8 & 31.3 & 62.4 \\
  \bottomrule
  \end{tabular}
}
\caption{\textbf{Impact of number of layers in the transformer projection head} on Something-Something v1. Time shift encoding is used for all runs. The performance begins to gradually saturate at four layers. The transformer projection head is only applied during pre-training, and is not used in downstream tasks.}
\label{tab:trn_ablation}
\end{table}

\noindent\textbf{2. Number of Pretraining Epochs.} 
We ablate the number of pretraining epochs when evaluated on SSv1, UCF101 and HMDB51. We observe in Table~\ref{tab:something_epoch} that pretraining for more epochs helps improve representation quality, as also observed by~\cite{chen2020simple}, and it saturates at ~500 epochs.

\begin{table}[H]\centering
\scalebox{0.8}{
  \begin{tabular}{cccc}
  \toprule
  Epochs & SSv1 & UCF101 & HMDB51 \\
  \midrule
  200 & 29.8 & 71.4 & 43.6 \\
  500 & 32.2 & 84.3 & 53.6 \\
  800 & 33.1 & 83.6 & 53.0 \\
  \bottomrule
  \end{tabular}
 }
\vspace{0.1cm}
\caption{\textbf{Impact of number of training epochs} on SSv1, UCF101 and HMDB51, using linear eval on frozen features.}
\label{tab:something_epoch}
\end{table}

\noindent\textbf{3. Results on SSv2.} 
We follow the same setup as Table 2 and study the impact of crop and time encodings when both the pretraining and target datasets are SSv2. Results are shown in Table~\ref{tab:compose_ssv2}. We observe a similar trend as in SSv1: encoding time outperforms the no encoding baseline, and composing time and crop encodings further improves performance.

\begin{table}[H]\centering
\scalebox{0.8}{
  \begin{tabular}{cccc}
  \toprule
  Enc. Crop & Enc. Time & Top-1 Acc. & Top-5 Acc. \\
  \midrule
  \xmark&\xmark&40.0&72.4\\
  \checkmark&\xmark&40.1&72.4\\
  \xmark&\checkmark&42.3&74.5\\
  \checkmark&\checkmark&\textbf{43.5}&\textbf{75.3}\\
  \bottomrule
  \end{tabular}
 }
\vspace{-0.1cm}
\caption{\textbf{Results on crop and time encodings on on SSv2} under a linear eval protocol. Trend is consistent with SSv1.}
\label{tab:compose_ssv2}
\end{table}

\noindent\textbf{4. Types of Action Classification.} In addition to results on SS, we also show results on standard action classification benchmarks UCF101 and HMDB51 under two settings -  using all frames and using only the first frame in Table~\ref{tab:ablation_ucf_hmdb}.
We only show results with time encoding - we find that unlike SSv1 and SSv2, using the crop encoding hurts the performance. This is interesting and we conjecture that the benefit of augmentation encoding depends on the downstream task at hand: for fine-grained tasks that require some level of spatial reasoning (\eg object localisation is needed to tell \textit{picking up} from \textit{putting down} in SSv1.), awareness of spatial augmentations is helpful; however for scene-level classification (\eg UCF101 and HMDB51) it might be beneficial to be invariant to those augmentations.

Table~\ref{tab:ablation_ucf_hmdb} shows a similar trend for encoding time as that on SSv1, improving over the baseline. The relative improvement is bigger for first frame classification vs using all frames, however for both cases, the relative improvement is smaller than on SSv1. Finally, we also report results on the Kinetics-400 dataset: Without encoding time shifts, the linear evaluation top-1 accuracy is 55.3\%. With encoding, the accuracy improves to 57.0\%. The relative improvement is similar to that on UCF-101 and HMDB-51, and smaller than on Something-Something. These are consistent with previous observations~\cite{s3dg_2017} that temporal information is more important for the Something-Something dataset.

\begin{table}[h]\centering
\scalebox{0.9}{
  \begin{tabular}{cccc}
  \toprule
  Input & Encode time & UCF & HMDB \\
  \midrule
  All frames & \xmark & 83.01 & 52.77\\
  All frames & \checkmark & \textbf{84.32} & \textbf{53.57}\\
  First frame & \xmark & 73.67 & 38.69\\
  First frame & \checkmark & \textbf{75.50} & \textbf{40.13} \\
\bottomrule
  \end{tabular}
\vspace{0.2cm}
\caption{\textbf{Effect of time encoding on UCF101~\cite{ucf101} and HMDB51~\cite{kuehne2011}} We show results for both early action classification (first frame) and regular action classification (all frames). We use frozen features: i.e. pretrained representations trained on Kinetics-400 are fixed and classified with a linear layer. Encoding time helps in both settings, albeit slightly more for early action classification.}
\label{tab:ablation_ucf_hmdb}
}
\end{table}

\noindent\textbf{5. Nearest neighbor retrieval.} We also validate our learned representations using the nearest neighbor retrieval benchmark. We follow the standard evaluation protocol~\cite{OPN,han2020memory}: For each query video in the test set, we retrieve its top $k$ nearest neighbors in the training set. A correct retrieval is deemed when any of the nearest neighbors belongs to the same category as the query video. We follow the linear evaluation procedure, and extract the visual representations from the visual encoders $f(\cot)$. For each video, we uniformly sample two windows of 32 frames and average their extracted representations. They are then $L_2$ normalized for retrieval. Following the standard protocol, we report results on the first split of UCF-101 and HMDB-51. As shown in Table~\ref{tab:knn_ucf} and~\ref{tab:knn_hmdb}, CATE significantly outperforms previous approaches in the video retrieval benchmark.

\begin{table}[H]
\centering
\resizebox{0.9\columnwidth}{!}{
\begin{tabular}{ c c c c c c}
\toprule
Method & top 1 & top 5 & top 10 & top 20 & top 50\\
\midrule
OPN~\cite{OPN} & 19.9 & 28.7 & 34.0 & 40.6 & 51.6\\
SpeedNet~\cite{benaim2020speednet}& 13.0 & 28.1 & 37.5 & 49.5 & 65.0\\
VCP~\cite{luo2020video} & 19.9 & 33.7 & 42.0 & 50.5 & 64.4\\ 
Temporal SSL~\cite{jenni2020video} & 26.1 & 48.5 & 59.1 & 69.6 & 82.8 \\
MemDPC$^\dagger$~\cite{han2020memory} & 40.2 & 63.2 & 71.9 & 78.6 & - \\
CATE & {\bf 54.9} & {\bf 68.3} & {\bf 75.1} & {\bf 82.3} & {\bf 89.9}\\
\bottomrule
\end{tabular}
}
\caption{Nearest neighbor retrieval evaluation on UCF-101 split 1. $\dagger$: with Flow}
\label{tab:knn_ucf}
\end{table}

\begin{table}[H]
\centering
\resizebox{0.9\columnwidth}{!}{
\begin{tabular}{ c c c c c c}
\toprule
Method & top 1 & top 5 & top 10 & top 20 & top 50\\
\midrule
VCP~\cite{luo2020video} & 6.7 & 21.3 & 32.7 & 49.2 & 73.3\\ 
MemDPC$^\dagger$~\cite{han2020memory} & 15.6 & 37.6 & 52.0 & 65.3 & -\\
CATE & {\bf 33.0} & {\bf 56.8} & {\bf 69.4} & {\bf 82.1} & {\bf 92.8}\\
\bottomrule
\end{tabular}
}
\caption{Nearest neighbor retrieval evaluation on HMDB-51 split 1. $\dagger$: with Flow}
\label{tab:knn_hmdb}
\end{table}

\noindent\textbf{6. Per-class breakdown on SSv1.} Table~\ref{tab:crop_per_class} shows the classes that benefit the most and the least when crop augmentation is encoded by CATE. As discussed in the main paper, the trend is consistent with results for time encoding, and indicates that crop encoding leads to representation that better captures spatial information.

\begin{figure*}[t]
\centering
\hspace{-2em}
\includegraphics[width=0.95\linewidth]{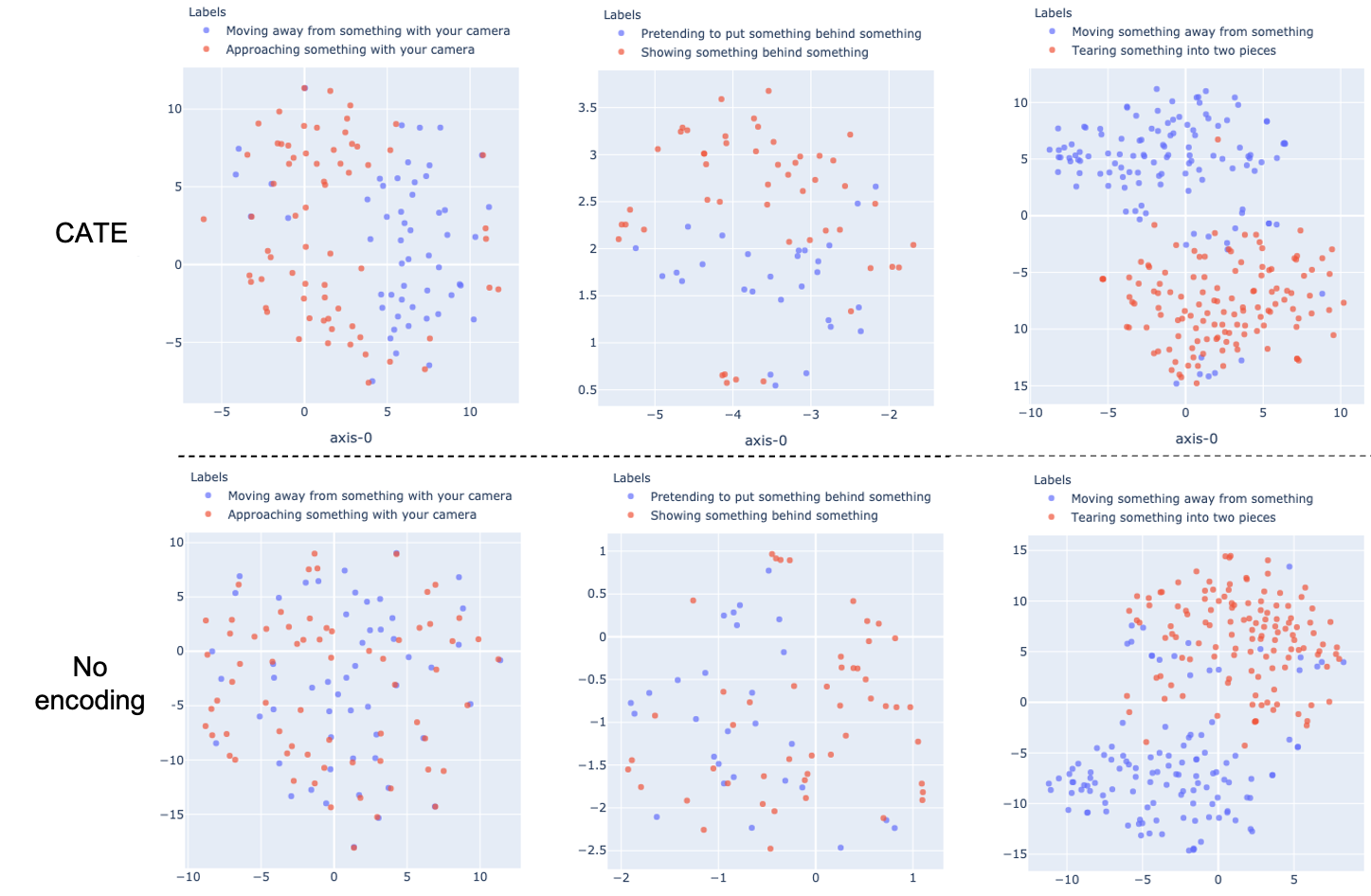}
\caption{t-SNE plots computed on the test split of SSv1 videos. The top row uses representations learned by CATE with time and crop encoding, the bottom rows uses representations learned without any augmentation encodings. For each column, we zoom into two categories which are colour-coded using red and blue. We observe that CATE in general leads to representations that better separate fine-grained action categories which are sensitive to temporal information (\eg \textit{moving away or approaching something with camera}) Best viewed in colour.
}
\label{fig:tsne} 
\end{figure*}

We further zoom into pairs of categories in Figure~\ref{fig:tsne} with t-SNE plots. We extract representations from the test split of SSv1, where the representational model from the top row is learned by CATE with crop and time encoding, while the bottom row is learned without augmentation encoding. We pick categories that are sensitive of temporal ordering, such as \textit{moving away} or \textit{approaching something with camera}, or \textit{pretending to put} or \textit{show something} behind something. We observe that CATE in general leads to representation that better separates these fine-grained actions, where no encoding leads to data points from different categories (red and blue in the figure) mix with each other.

\begin{table}[h]\centering
\scalebox{0.8}{
  \begin{tabular}{cc}
  \toprule
  Label & $\Delta\textrm{AP}$\\
  \midrule
  \rowcolor{aliceblue}Lifting something up completely, then letting it drop down & 13.5\\
  \rowcolor{aliceblue}Pulling something from right to left & 13.2\\
  \rowcolor{aliceblue}Moving something and something away from each other & 13.2\\
  \rowcolor{aliceblue}Dropping something in front of something & 12.6\\
  \rowcolor{aliceblue}Moving something down & 12.2\\
  \rowcolor{mistyrose}Pretending to sprinkle air onto something & -7.0\\
  \rowcolor{mistyrose}Folding something & -8.6\\
  \rowcolor{mistyrose}Pretending or failing to wipe something off of something & -10.0\\
  \rowcolor{mistyrose}Moving away from something with your camera & -11.6\\
  \bottomrule
  \end{tabular}
 }

\caption{Classes that benefit the most and the least with \textbf{crop encoding} on SSv1. We sort the classes by their differences on Average Precision.}
\label{tab:crop_per_class}
\end{table}

\subsection{Results on CLEVR and DSprites}
Additionally, we further study the impact of crop encoding by using two image benchmarks that explicitly require spatial reasoning. The first dataset is CLEVR~\cite{clevr} with 70,000 training and 15,000 validation images. It is a diagnostic dataset which contains multiple objects of diverse shape and location configurations. We follow the setup used by~\cite{zhai2019visual} and evaluate on two tasks: \textbf{Count} which requires counting the total number of objects, and \textbf{Dist} which requires predicting the depth of the closest object to the camera, where the depth is bucketed into 6 bins. Both tasks are formulated as classification tasks. The second dataset is DSprites~\cite{dsprites17} which contains a single object floating around in an image, with various shape, scale, orientation and location. We use the \textbf{Location} task which requires predicting the $(x,y)$ center location of the object. The $x$ and $y$ coordinates are bucketed into 16 bins each. We report the geometric mean of classification accuracy on the bucketed $x$ and $y$ coordinates.

For both benchmarks, we train \model using the same setups as we did with videos, except that the visual encoder is now a 2D ResNet-50, and the learning rate is reduced by 5x. We pretrain and evaluate on the datasets themselves.

\begin{table}[H]\centering
\scalebox{0.8}{
  \begin{tabular}{cccccc}
  \toprule
  Crop Enc. & CLEVR-Count~\cite{clevr} & CLEVR-Dist~\cite{clevr} & DSprites~\cite{dsprites17}\\
  \midrule
   & 65.3 & 64.3 & 28.1\\
  \checkmark & \textbf{68.8} & \textbf{66.9} & \textbf{38.8}\\
  \bottomrule
  \end{tabular}
 }

\caption{Ablation of crop encoding on downstream tasks that require spatial reasoning, such as counting the number of objects, or localising objects in bucketed x, y coordinates.}
\label{tab:image_encoding}
\end{table}

The linear evaluation performance is shown in Table~\ref{tab:image_encoding}. We observe that encoding crop improves the transfer learning performance on all three tasks that require spatial reasoning, which further validates our conjecture.

\end{document}